\documentclass{l4dc2025}

\usepackage{times}
\usepackage{subfiles}
\usepackage{array}
\usepackage{adjustbox}
\usepackage{paralist}
\usepackage{wrapfig}
\usepackage{float}

\newtheorem{assumption_}{\textbf{Assumption}}
\newtheorem{lemma_}{\textbf{Lemma}}

\newtheorem{theorem_}{\textbf{Theorem}}
\newtheorem{definition_}{\textbf{Definition}}

\DeclareMathOperator*{\argmin}{arg\,min}

\usepackage{soul}

\makeatletter
\newcommand*{\centerfloat}{%
  \parindent \z@
  \leftskip \z@ \@plus 1fil \@minus \textwidth
  \rightskip\leftskip
  \parfillskip \z@skip}
\makeatother

\usepackage{enumitem}
\setlist{nolistsep}

\title[Symmetries-enhanced MARL]{Symmetries-enhanced Multi-Agent Reinforcement Learning}

\author{%
 \Name{Nikolaos Bousias\textsuperscript{1}} \Email{nbousias@seas.upenn.edu}\\
 \Name{Stefanos Pertigkiozoglou\textsuperscript{1}} \Email{pstefano@seas.upenn.edu}\\
 \Name{Kostas Daniilidis\textsuperscript{1,2}} \Email{kostas@cis.upenn.edu}\\
 \Name{George Pappas\textsuperscript{1}} \Email{pappasg@seas.upenn.edu}\\
 \addr \textsuperscript{1}GRASP Lab, University of Pennsylvania, Philadelphia, PA,
 \addr \textsuperscript{2}Archimedes, Athena RC
}

\begin{document}
\maketitle

\begin{abstract}%
Multi-agent reinforcement learning has emerged as a powerful framework for enabling agents to learn complex, coordinated behaviors but faces persistent challenges regarding its generalization, scalability and sample efficiency. Recent advancements have sought to alleviate those issues by embedding intrinsic symmetries of the systems in the policy. Yet, most dynamical systems exhibit little to no symmetries to exploit. This paper presents a novel framework for embedding extrinsic symmetries in multi-agent system dynamics that enables the use of symmetry-enhanced methods to address systems with insufficient intrinsic symmetries, expanding the scope of equivariant learning to a wide variety of MARL problems. Central to our framework is the Group Equivariant Graphormer, a group-modular architecture specifically designed for distributed swarming tasks. Extensive experiments on a swarm of symmetry-breaking quadrotors validate the effectiveness of our approach, showcasing its potential for improved generalization and zero-shot scalability. Our method achieves significant reductions in collision rates and enhances task success rates across a diverse range of scenarios and varying swarm sizes.
\end{abstract}


\section{Introduction}

The study of multi-agent dynamical systems, has garnered significant attention due to its wide-ranging applications in areas like autonomous navigation, environmental monitoring, target tracking, collaborative manipulation etc. Since obtaining large-scale datasets of expert demonstrations is often impractical, multi-agent reinforcement learning (MARL) has emerged as a powerful framework for enabling agents to learn complex, coordinated behaviors. However, the inherently high-dimensional and decentralized nature of multi-agent systems poses significant challenges, particularly in terms of scalability, generalization, and data efficiency. To enhance generalization, synthetic samples can be generated by applying symmetry transformations to the original data, increasing diversity without additional sampling (e.g. \cite{NEURIPS2020_e615c82a,kostrikov2021imageaugmentationneedregularizing}). Data augmentation, though, increases computation time and offers no generalization assurances. Outside of data augmentation, one promising avenue to address these challenges lies in leveraging the symmetries of dynamical systems as a policy inductive bias. This paper focuses on extrinsic symmetry exploitation in cooperative-competitive MARL problems.

From an RL perspective, the existence of symmetries morphs into equivalence of state-action pairs under a group transformation, meaning that the policy only needs to learn one mapping for each equivalence class, rather than separately learning the same behavior for all symmetrically related pairs \cite{Ravindran2001SymmetriesAM,Zinkevich2001SymmetryIM}. The presence of symmetries, therefore, reduces the hypothesis space and improves sample efficiency and generalization, as the policy learned for one pair generalizes to all equivalent pairs. Symmetry-aware methods can effectively operate on a reduced state-action space, corresponding to the equivalence classes rather than individual pairs (\cite{Yu2022EquivariantRL,sonmez2024exploitingsymmetrydynamicsmodelbased,10054413,equivariant_downwash}). Though working with representations of the equivalence classes benefits from off-the-shelf neural architectures to parametrize the policy, it may restrict its expressivity, thus motivating the design of neural architectures that inherently respect the symmetries as an inductive bias (\cite{mondal2020groupequivariantdeepreinforcement,vanderpol2020plannableapproximationsmdphomomorphisms,NEURIPS2020_2be5f9c2,zhu2022sampleefficientgrasplearning,simm2020symmetryawareactorcritic3dmolecular,pmlr-v164-wang22j,wang2020policylearningse3action,nguyen2023equivariant}). These approaches are only suitable for single agent or centralized multi-agent controllers and the neural networks used are tailored to the specificities of the system.

Omnipresent in homogeneous multi-agent systems is permutation equivariance, i.e. dynamics and reward functions are identical across robots, so indexing is interchangeable. This discrete $S_n$ symmetry, leading to weight sharing among agents, is the basis of the representation power of GNNs (\cite{jiang2020graphconvolutionalreinforcementlearning,tzes2023graph}) and directly limits the required sample complexity for training the distributed policy. Permutation symmetries have been explored in MARL via policy sharing (\cite{hao2023boosting,chen2022communicationefficient,NIPS2017_c2ba1bc5,Pol2016Coordinated,pmlr-v235-lin24m}), permutation equivariant critics (\cite{Almasan_2022,liu2019picpermutationinvariantcritic,rashid2020monotonicvaluefunctionfactorisation,liu2019picpermutationinvariantcritic}) or mean field approximation (\cite{pmlr-v80-yang18d}). These approaches exploit the homogeneity structure but ignore the geometry properties of the problem. From a geometric standpoint, few symmetries in MARL have been explored, namely discrete $C_n$ rotation symmetries \cite{vanderpol2022multiagentmdphomomorphicnetworks} and $E(3)$ \cite{pmlr-v202-chen23i, chen2023rm, mcclellan2024boostingsampleefficiencygeneralization}. The existing works commonly employ equivariance-by-construction policies based on equivariant operations on message-passing networks (\cite{pmlr-v139-satorras21a}) that assume $E(n)$ symmetry of the problem and, thus, are not transferable to MARL problems with different symmetries. Inspired by \cite{wang2023the,HAMPSEY2023132}, we propose a simple, yet modular, Group Equivariant Graphormer that can be adapted for a variety of symmetries in differently structured MARL problems via canonicalizing group actions on tensorial graph features. Existing literature focuses on exploiting existing symmetries in the systems, whereas the current paper extends to systems with partial or broken symmetries by embedding them in associated symmetry-enhanced systems to include desirable equivariant properties  
 \vspace{-3mm}
\subsection{Contributions}
The aforementioned literature\footnote{Supplemental material can be found in Appendices I-IV} utilizing symmetries as inductive biases in policy learning, which, however, presumes that the system exhibits said symmetries. In practice, most dynamical systems exhibit little to no symmetries. To the best of our knowledge, this is the first paper that attempts to construct a MARL framework that leverages symmetries in policies, even if the system is not endowed with them. Our contributions are outlined below:
\begin{itemize}
\item A formalization of the symmetry properties of multi-robot dynamical systems and the conditions under which optimal policies are equivariant functions to be approximated by equivariant networks. These conditions showcase the restrictive nature of current equivariant RL methods to systems with explicit symmetries shared by the task specification.

\item A methodology for embedding extrinsic symmetries in systems where intrinsic symmetries are insufficient or broken, thus broadening the applicability of equivariant learning frameworks to any multi-agent dynamical system.

\item The introduction of a Group Equivariant Graphormer architecture tailored to distributed swarming tasks. This network is by design modular in terms of the symmetry considered (any group compatible with the manifold of the dynamics, discrete or continuous), in contrast to the vast majority of networks in the literature. 

\end{itemize}
By embedding extrinsic symmetries, our approach demonstrates enhanced learning efficiency and generalization capabilities, paving the way for broader adoption of symmetry-aware methods in MARL and control. We subsequently validate its efficacy through extensive experimentation on a swarm of $SE(3)$ symmetry-breaking quadrotors that showcases the effectiveness of our scheme regarding scalability and generalization.

\section{Preliminaries}

\subsection{Group Theory \& Equivariant Functions}
A group $(G,\cdot)$ is a set $G$ equipped with an operator $\cdot:G\times G \rightarrow G$ that satisfies the properties of \emph{Identity:} $\exists e \in G$ such that $e \cdot g = g \cdot e = e$; \emph{Associativity:} $\forall g,h,f \in G,\, g\cdot (h \cdot f) = (g \cdot h) \cdot f$; \emph{Inverse:} $\forall g \in G,\, \exists g^{-1}$ such that $g^{-1} \cdot g = g \cdot g^{-1} = e$. Additional to its structure we can define the way that the group elements act on a space $X$ via a group action:
\begin{definition_}\label{definitioin:group_action}
A map $\phi_g: X\to X$ is called an action of group element $g\in G$ on $X$ if for $e$ is the identity element $\phi_e(x)=x$ for all $x\in X$ and $\phi_g\circ \phi_h=\phi_{g\cdot h}$ for all $g,h\in G$.
\end{definition_}
Note here that a group action on a given space $X$ allow us to group different elements of $X$ in sets of orbits. More precisely given a group action $\phi_*$ an orbit of a element $x\in X$ is the set $\mathcal{O}_x^{\phi_*}=\left\{\phi_g(x)|g\in G\right\}$. 
\noindent In many application we require functions that respect the structure of a  group acting on their domain and codomain. 
We refer to these functions as equivariant and we formally define them as follow:
\begin{definition_}
Given a group $G$ and corresponding group actions $\phi_g:X\to X$, $\psi_g:X\to X$ for $g\in G$  a function $f:X\to Y$ is said to be $(G,\phi_*,\psi_*)$ equivariant if and only if: 
\begin{equation}
    \psi_g\left[f(x)\right]=f\left(\phi_g[x]\right)\quad \forall x\in X, g\in G
\end{equation}
\end{definition_}

\subsection{Notation}
    Let $\mathcal{X}$ be a smooth manifold and $T_x\mathcal{X}$ the tangent space at an arbitrary $x\in \mathcal{X}$. A smooth vector field is a smooth map $f:\mathcal{X}\rightarrow T\mathcal{X}$ with $f(x)\in T_x\mathcal{X}$. The set of smooth vector fields on a manifold $\mathcal{X}$, denoted $\mathfrak{X}(\mathcal{X})$, is a linear infinite dimensional vector space. Let $G$ be a d-dimension real Lie group, with identity element $e$. 
For a smooth manifold $\mathcal{X}$ and Lie group $G$, the left natural action $\phi:G \times \mathcal{X} \rightarrow \mathcal{X}$ satisfies $\phi(e,x) = x,\forall x\in \mathcal{X}$ and $\phi(\hat{g},\phi(g,x))=\phi(\hat{g}\cdot g , x),\forall g.\hat{g}\in G, x\in \mathcal{X}$, thus inducing families of smooth diffeomorphisms $\phi_g(x):=\phi(g,x)$. A group action is \textit{free} if $\forall x \in \mathcal{X},\phi(g,x)=x \Leftrightarrow g=e $ and \textit{transitive} if $\forall x,y \in \mathcal{X} \,,\,\exists g\in G$ such that $\phi(g,x)=y$ (i.e. the nonlinear smooth projections $\phi_x(g):=\phi(g,x)$ are surjective). 
    A \textit{homogeneous} space is a smooth manifold $\mathcal{X}$ that admits a transitive group action $\phi: G \times \mathcal{X} \rightarrow \mathcal{X}$ and the Lie group $G$ is, then, called the \textit{symmetry} of $\mathcal{X}$. The group torsor $\mathfrak{G}$ of a Lie group $G$ is defined as the underlying manifold of $G$ without the group structure, allowing for identification of the torsor elements by the group elements, denoted $\chi\in \mathfrak{G} \simeq g \in G$, and inheriting the free and transitive group action $\phi$ induced by the group operator, i.e. for $\hat{g}\in G$ and $\chi\in \mathfrak{G} \simeq g \in G$ it stands that $\phi(\hat{g},\chi) \simeq \hat{g}\cdot g$. Crucially, a manifold that serves as a torsor for multiple Lie groups may admit multiple symmetries.

\subsection{Problem Statement}\label{problem_statement}
Consider a homogeneous multi-robot dynamical system, comprising of $N$ autonomous robots indexed $i\in \{1,\dots,N\} \equiv I_N$. Let $\mathcal{X}$ be a smooth manifold and $\mathcal{U}$ a finite dimensional input space. The agents are described by dynamics:
\begin{align}
    \dot{x}_i(t) = f(x_i(t),u_i(t)),\quad x_i(0)=x_{i}^0,\forall i\in I_N
\end{align}
where $u_i\in \mathcal{U},x_i\in\mathcal{X}$ and $f:\mathcal{X}\times \mathcal{U} \rightarrow \mathfrak{X}(\mathcal{X})$ a linear morphism. Consider a Lie group $G$ and a smooth transitive group action $\phi:G \times \mathcal{X}\rightarrow \mathcal{X}$. 

\noindent \textbf{Graph Representation of Multi-robot Systems}\label{intro:graph_representation}: We assume that the robots are equipped with sensing/communication capabilities with a range $\hat{\epsilon}$. Let $\mathcal{N}_i:= \{j\in I_N \setminus\{i\} \,:\, d(x_i,x_j)\leq \hat{\epsilon} \}\,,\, \forall i \in I_N$ denote the neighbourhoods, thus giving rise to a graph representation of the system $\mathcal{G}_t=\{V
_{\mathcal{G}_t},\mathcal{E}_{\mathcal{G}_t}\}$, with nodes $V
_{\mathcal{G}_t}=\{x_i(t)\,,\, i\in I_N\}$ and edges $\mathcal{E}_{\mathcal{G}_t}=\{ (i,j)\,,\, : \forall i\in I_N,j\in \mathcal{N}_i \}$. Information is propagated over the graph structure, with each robot receiving local observations of the system, denoted $o_i(t)=\{x_j(t)\,,\, j\in \mathcal{N}_i\}\,,\, i\in I_N$. We denote $x(t)=[\oplus_{\forall i\in I_N} x_i(t)]\in \cup_{i\in I_N}\mathcal{X}$ and $u(t)=[\oplus_{\forall i\in I_N} u_i(t)]\in \cup_{i\in I_N}\mathcal{U}$ the centralized state and action of the multi-robot system respectively. Assuming that a submanifold $\Bar{X} \subseteq\mathcal{X}$ forms the torsor $\mathfrak{G}$ of the Lie group $G$ and that $\mathcal{X}\setminus \Bar{X}$ is compatible with $G$, then every robot inherits a group representation element $g_i\,,\forall i\in I_N$ and the node attributes of the graph become $V_{\mathcal{G}_t}=\{(g_i(t),x_i(t))\,,\, i\in I_N\}$.
 
\noindent \textbf{Problem statement}: Given a swarm of $N$ robots of known dynamics, the trajectories $\{x_i(t)\,,\, i\in I_N\},t\in \mathbb{R}^{+}$ evolve in a complex environment, known or estimated continuously through onboard sensors, with obstacles represented as a point cloud set $\mathcal{O}\in \mathbb{R}^{3 \times d}$. Assuming a metric function denoted $\mathcal{L}: \cup_{i\in I_N} \mathcal{X} \times \cup_{i\in I_N} \mathcal{U} \rightarrow \mathbb{R}$ that codifies the specific task (e.g. distance for navigation, alignment for flocking), we formally define the multi-agent geometric optimization problem
\vspace{-2mm}\begin{align}
\label{eq:prob}
    u^{*}_{0:T,I_N}=& \argmin_{u_{0:T}} \int_{0}^T \mathcal{L}(\oplus_{I_N}x_i(\tau), \oplus_{I_N}u_i(\tau)) d\tau \nonumber\\
    s.t.\quad& \dot{x}_i(t)=f(x_i(t),u_i(t))\in \mathfrak{X}(\mathcal{X})\,,\, i\in I_N \nonumber\\
    &u_i(t) = \pi(o_i(t) \cup x_i(t);\mathcal{O}) \in \mathcal{U}
\vspace{-2mm}\end{align}
Problem (\ref{eq:prob}) amounts to $N$ robots learning the distributed control policy $\pi_\theta: \cup_{\mathcal{N}_i \cup \{i\}} \mathcal{X} \times \mathbb{R}^{3 \times d} \rightarrow \mathcal{U} $ using only local observations and knowledge of the environment. This problem can be recast as a distributed MARL problem by maximizing a reward function instead of minimizing a loss function, with robot dynamics described by a transition function, forming an MDP. For clarity reasons we opt to use the optimal control jargon.

\section{Exploiting Symmetries of Dynamical Systems}\label{section:EquivariantProblem}

 In this section we show that if Problem \ref{eq:prob} exhibits some structural symmetries then the optimal control policy needs to exhibit them as well, leading to sample efficient learning. 
\begin{definition_}
    \label{eq:G-equivariant_problem}
    Consider a real Lie Group $G$, a smooth manifold with group properties satisfying differentiability of group operations. Problem \ref{eq:prob} is $G$-equivariant if, for transitive actions induced by elements of the group $G$ on a vector field $\mathcal{X}$, $\phi:\mathcal{X}\times G \rightarrow \mathcal{X}$ and $ \psi:\mathcal{U}\times G \rightarrow \mathcal{U}$, it satisfies the following properties:
    \begin{enumerate}
        \item \label{eq:G-equivariant_problem1} The objective function is type-0 equivariant (invariant function), i.e. \vspace{-2mm}\begin{align*}
            \mathcal{L}(\oplus_{I_N}x_i(t), \oplus_{I_N}u_i(t) )=\mathcal{L}(\oplus_{I_N}\phi_g(x_i(t) ), \oplus_{I_N}\psi_g(u_i(t) ) ),\,\forall g \in G  \end{align*}\vspace{-7mm}
        \item \label{eq:G-equivariant_problem2} The robot dynamics are equivariant w.r.t. elements of $G$, i.e. \vspace{-2mm}\begin{align*} d\phi_gf(x_i(t),u_i(t)) = f(\phi_g(x_i(t)), \psi_g(u_i(t)),\,\forall g \in G \end{align*}\vspace{-2mm} where $d\phi: \mathfrak{X}(\mathcal{X}) \times G\rightarrow \mathfrak{X}(\mathcal{X})$ the differential of the diffeomorphism defining the symmetry.
    \end{enumerate}
\end{definition_}
In RL terminology, this is equivalent to requiring the transition, reward, and observations functions to be invariant under group actions, giving rise to the G-invariant Markov Games.
\begin{assumption_}\label{assumption:G-invarinat neighborhood}
    The neighborhood structure $\mathcal{N}_i,\forall i\in I_N$ is G-invariant.
\end{assumption_}
\begin{theorem_}\label{theorem:equivariant_optimal_control}
    The optimal control policy $\pi^{*}(x_i(t)\cup o_i(t))$ for the G-equivariant multi-robot problem, i.e. equation (\ref{eq:prob}) with definition (\ref{eq:G-equivariant_problem}), is equivariant under group actions from elements of $G$. (Proof is appended in Appendix I  of the supplemental material\footref{supplemental}.)
\end{theorem_}
Theorem \ref{theorem:equivariant_optimal_control} offers a way to shrink the hypothesis class of the learned controller to that of all G-equivariant functions. This inductive bias usually translates to fewer parameters required and greater sampling efficiency as multiple states $x(t)$ of the G-equivariant problem identified  via the actions of the group $G$, are essentially equivalent, and therefore any G-equivariant policy trained on a dataset $D=\{x^1,x^2,\dots \}$ would generalize to the dataset $\hat{D} = \cup_{\forall g\in G} \{\phi_g(x^1),\phi_g(x^2),\dots\}$ where $D\subseteq \hat{D} \subseteq \cup_{I_N}\mathcal{X}$. Leveraging symmetries is particularly important in multi-agent RL as it is a sampling inefficient learning process.


\section{When symmetries are broken}\label{section:ArtificialEquivariance}

Section \ref{section:EquivariantProblem} demonstrated why symmetries of dynamical systems are useful. Even though most tasks exhibit geometric symmetries (condition 1 of Definition \ref{eq:G-equivariant_problem}), most dynamical systems have little to no symmetries to exploit (condition 2), rendering Theorem \ref{theorem:equivariant_optimal_control} void. To circumvent this issue, we propose embedding the non-equivariant system into an extended dynamical system that is G-equivariant, so that Theorem \ref{theorem:equivariant_optimal_control} stands. The extended learned policy is G-equivariant, and optimal action for the original non-equivariant dynamical system is recovered via the symmetry-breaking projective function that guarantees equivalence between the two associated dynamical systems. This way we can provide structure in the learned policy, even if the original dynamical system does not support it. A consequence of this framework is that the structure of the policy is solely constrained by the symmetries that the reward function admits, meaning that any problem can be recast as a G-equivariant problem as long as condition \ref{eq:G-equivariant_problem1} holds and the Lie group G is compatible with the manifold $\mathcal{X}$. 
Consider the push-forward smooth map $d_{*}\phi : \mathfrak{X}(\mathcal{X}) \times G\rightarrow \mathfrak{X}(\mathcal{X})$ defined as $d_{*}\phi_g f(x,u):= d\phi_g f (\phi_{g^{-1}}(x),u)$, naturally induced by the diffeomorphism $\phi: G\times \mathcal{X} \rightarrow \mathcal{X}$.
\begin{lemma_}\label{lemma:push_forward}
     The push-forward function $d_{*}\phi$ is a well defined linear group action (automorphism) on the infinite dimensional vector field $\mathfrak{X}(\mathcal{X})$. (Proof in Appendix I of supplemental material\footref{supplemental}.)
\end{lemma_}
\begin{theorem_}\label{theorem:equivariant_input_extension}
     Let $f:\mathcal{X}\times \mathcal{U} \rightarrow \mathfrak{X}(\mathcal{X})$ denote a control affine dynamical system on a smooth manifold $\mathcal{X}$ that admits a transitive group action from a compatible Lie group $G$, over an input vector space $\mathcal{U}$. For the extended input vector space $\hat{\mathcal{U}} := span\{d_{*}\phi_{g}f(x,u)\,|\,  u \in \mathcal{U},g \in G\}$, the associated system dynamics  $F:\hat{\mathcal{U}} \times \mathcal{X} \rightarrow \mathfrak{X}(\mathcal{X})$ are equivariant with respect to actions induced by elements of the Lie group $G$. (Proof in Appendix I of supplemental material\footref{supplemental}.)
\end{theorem_}
Leveraging theorem \ref{theorem:equivariant_input_extension}, any control affine dynamical system can be embedded in an associated equivariant dynamical system, by extending the input to include the closure of the image of the dynamics under the pushforward operator. The associated dynamics then are $F(x,\hat{u}) := \sum_{j=1}^K d_{*}\phi_{g_j}f(x,u_j)$, for $g_i\in G,u_i\in \mathcal{U},K\in\mathbb{N}, x\in \mathfrak{X}(\mathcal{X}),\hat{u}\in \hat{\mathcal{U}}$. A limiting factor is that, even if the original input space $\mathcal{U}$ is finite dimensional, the extended input vector space $\hat{\mathcal{U}}\subset\mathfrak{X}(\mathcal{X})$ may be infinite dimensional. In practice, its dimension depends on the complexity of the dynamical system and the selected symmetry group. 
\begin{remark}(\cite{Mahony2020EquivariantST}\label{remark:2}
    The trajectory of a control affine dynamical system $f:\mathcal{X}\times \mathcal{U} \rightarrow \mathfrak{X}(\mathcal{X})$ on a smooth manifold $\mathcal{X}$ is identical to the trajectory of its associated system dynamics $F:\hat{\mathcal{U}} \times \mathcal{X} \rightarrow \mathfrak{X}(\mathcal{X})$, defined in theorem (\ref{theorem:equivariant_input_extension}), if the extended control input is constrained in $\mathcal{U}$, i.e. for some $u(t) \in \mathcal{U}$ $F(x(t),u(t))=f(x(t),u(t))$.
\end{remark}
We can, then, reformulate problem (\ref{eq:prob}) to incorporate artificial symmetries even if the dynamical system does not satisfy the condition \ref{eq:G-equivariant_problem2} of definition \ref{eq:G-equivariant_problem} (only condition \ref{eq:G-equivariant_problem1}) as 
\vspace*{-2mm}
\begin{align}\label{eq:equivariant_problem}
    u^{*}_{0:T,I_N}=& h_{\mathcal{U}}(\argmin_{\hat{u}_{0:T}\in \hat{\mathcal{U}}} \int_{0}^T \mathcal{L}(\oplus_{I_N}x_i(\tau) , \oplus_{I_N} h_\mathcal{U}(\hat{u}_i(\tau))) d\tau) \nonumber\\
    s.t.\quad& \dot{x}_i(t)=F(x_i(t),h_\mathcal{U}(\hat{u}_i(t)))\in \mathfrak{X}(\mathcal{X})\,,\, i\in I_N \nonumber\\
    &\hat{u}_i(t) = \hat{\pi}(o_i(t) \cup x_i(t);\mathcal{O}) \in \hat{\mathcal{U}}
\vspace{-2mm}\end{align}
where $h_{\mathcal{U}}:\hat{\mathcal{U}} \rightarrow \mathcal{U}$ a smooth idempotent surjective morphism constraining the extended input to the original input vector space $\mathcal{U} \subset \hat{\mathcal{U}}$.
\begin{proposition}\label{theorem:equivariant_problem}
    Problems (\ref{eq:prob}) and (\ref{eq:equivariant_problem}) are equivalent and, if $\mathcal{L}:\mathcal{X} \rightarrow \mathbb{R}$ is G-invariant, there exists an optimal lifted $\hat{\pi}^{*}:\mathcal{X}^{N} \rightarrow \hat{\mathcal{U}}$ that is G-equivariant. (Proof in supplemental material\footref{supplemental}.)
\end{proposition}
From Proposition \ref{theorem:equivariant_problem} there exists a G-equivariant extended optimal policy $\hat{\pi}^{*}: \mathcal{X}^{|\mathcal{N}_i|+1} \rightarrow \hat{\mathcal{U}}$ such that the optimal policy can be decomposed as $\pi^{*}(x_i(t),o_i(t))= h_\mathcal{U} \circ \hat{\pi}^{*} (x_i(t),o_i(t))$. Assuming that the extended input space is finite-dimensional, the universal approximation theorem holds, meaning that the equivariant policy for the associated system can be $\epsilon$-approximated by any G-equivariant neural network $\hat{\pi}_\theta$. Unfortunately, finding the associated equivariant dynamical system is a difficult process, even for systems with simple dynamics, so $h_\mathcal{U}$ is unknown and must also be learned by a neural network $h_\theta$. Without the equivariant analogue system $F$, the policy $\pi_\theta(x_i,o_i)= h_\theta \circ \hat{\pi}_\theta(x_i,o_i) \in \mathcal{U}$ is fitted using data drawn from the original non-equivariant system $f$, whose trajectories identify with $F$, thus there is no guarantee that $\hat{\pi}_\theta \rightarrow \hat{\pi}^{*}$. Still, the existence of $F$ alone guarantees that the compositional policy $\pi_\theta$ is valid, i.e. is a universal approximator of $\pi^{*}$. We demonstrate experimentally in Section \ref{section:experiments} the benefits in generalization and scalability of parametrizing the policy as a composition of a G-equivariant and a non-equivariant neural network.


\section{Equivariant Graphormer}\label{section:LieGraphormer}
To solve the Geometric Swarming problem, we must learn the distributed equivariant policy $\hat{u}_i(t)=\hat{\pi}(o_i \cup x_i;\mathcal{O}) \,,\, i\in I_N$, as described in Section \ref{section:ArtificialEquivariance}. Generally deep learning models require specialized architectures to ensure the satisfaction of the equivariant constraints \cite{fuchs2020se3transformers}. Such architectures can result in more challenging optimization \citep{pertigkiozoglou2024EqRelax} that can complicate their integration with standard RL techniques. To address these challenges we proposed to leverage the structure of our input graph representation and achieve equivariance through group canonicalization, described in Section \ref{sec:groupcanon}. This technique doesn't impose any additional constraints to the model architecture and thus it allows us to easily extend pre-existing arcitectures used by previous work to learn distributed swarming policies.  To that end, in Section \ref{sec:equivgraphTransformer} we describe an equivariant extension of the Graph Transformer \cite{müller2023attending}. 
\subsection{Group Canonicalization}\label{sec:groupcanon}
Given a group $G$ acting on space $X$ through action $\phi_g$ we can define an extended group action $\phi'_g$ on space $G\times X$ as $\phi'_g[(p,x)]=(g\cdot p,\phi_g[x])$, where $\forall g,p\in G, x\in X$.
Since $\phi_g$ is an action of group $G$, $\phi'_g$ satisfies the properties of definition \ref{definitioin:group_action} and, thus, it is also an action.
Assuming an additional space $Y$ with corresponding group action $\psi_g:G\times Y\to Y$, we can show the following:
\begin{lemma_}\label{lem:canonic}
A function $f:G\times X\to Y$ satisfies the equivariant constraint $f(\phi'_g[p])=\psi_g[f(p)]$, $\forall g\in G$ and $p\in G\times X$, if and only if, for $h:X\to Y$, it can be written as a composition:
\begin{align*}
    f(g,x)=\psi_g[h(\phi_{g^{-1}}[x])],\quad \forall g\in G, x\in X
\end{align*}
with $h(x)=f(e,x)$ for all $x\in X$ and $e$ being the identity element of $G$. (Proof in Appendix I of the supplementary material\footref{supplemental}.)
\end{lemma_}
Notice that in Lemma \ref{lem:canonic}, $h$ is a function from $X$ to $Y$ without any additional constraints. This implies that we can use any general function approximator (e.g. MLP, Transformer) to approximate an equivariant function $f:G\times X\to Y$, by only applying the appropriate input-output transformations. We will use this observation to extend the baseline non-equivariant models to be equivariant with minimal changes to the underlying architecture.
\subsection{Equivariant Graph Transformer}\label{sec:equivgraphTransformer}
 Given a feature augmented graph representation $(V,E,F)$ with a finite set of nodes $V$, a finite set of edges $E(G)\subset \{(u,v)|u,v\in V\}$ and a set of per-node features $F=\{f_v\in X|v\in V\}$, a graph transformer sequentially updates the nodes features using a local attention layer to aggregate information from neighboring nodes. Specifically the $l^{th}$ update layer for node $v\in V$ is a function $M_v: X^{(l)}\to X^{(l+1)}$ defined as $f_v^{(l+1)}\leftarrow M_v(F^{(l)})=\mu(f_v^{(l)}+\mathrm{attn}(F^{(l)})_v)$, where $\mu$ corresponds to a fully-connected feedforward network, and $\mathrm{attn}$ corresponds to the local attention layer: 
\begin{align*}
    \mathrm{attn}(F^{(l)})_v=\sum_{p\in \mathcal{N}_v}\frac{\mathrm{exp} \left( {f_v^{(l)}}^T W_Q^T W_Kf_p^{(l)} \right) }{\sum_{p\in \mathcal{N}_v}\mathrm{exp} \left({f_v^{(l)}}^T W_Q^T W_K f_p^{(l)} \right)} \left(W_V f_p^{(l)} \right)
\end{align*}
with $\mathcal{N}_v$ being the set of neighbors for nodes $v$.
As discussed in Section \ref{intro:graph_representation}, the input graph representation is endowed an additional structure that allows for an simple extention of the graph transformer to be equivariant. Specifically, each node $v\in V$ additional to a feature $f_v\in X$ describing each state is also equipped with a local frame $g_v\in G$. This means that the input graph is represented  as $(V,E,F_\mathrm{tens})$, with $F_\mathrm{tens}=\{(g_v,f_v)\in G\times X|v\in V\}$ being a set of "tensorial" features that are described by their local frame $g_v\in G$ along with their feature value $f_v$. For such a feature $(g_v,f_v)$ expressed in local frame $g_v\in G$ we can compute the equivalent feature expressed in a new frame $g_n\in G$ by applying the action $\phi'_{g_ng_v^{-1}}[(g_v,f_v)]=(g_n,\phi_{g_ng_v^{-1}}[f_v])$. This structure allows us to leverage the results of Lemma \ref{lem:canonic} and define an equivariant update rule $M^\mathrm{eq}_v:G\times X^{(l)}\to X^{(l+1)} $ for the features of node $v\in V$ as follows:
\vspace{-2mm}\begin{align*}
    f_v^{(l+1)}\leftarrow M^\mathrm{eq}_v(g_v,F^{(l)})=\phi^{(l+1)}_{g_v}\left[M_v\left(\phi^{(l)}_{g_v^{-1}}\left[F^{(l)}\right]\right)\right]
\vspace{-2mm}\end{align*}
with $\phi^{(l)}, \phi^{(l+1)}$ being actions of group $G$ on the corresponding input/ouput feature space $X^{(l)}, X^{(l+1)}$. By sequentially composing the equivariant update layers we define an edge-preserving isomorphism, end-to-end Group Equivariant Graphomer, which is used to learn the equivariant policy $\hat{\pi}$. It is easy to see that the Group Equivariant Graphormer is permutation equivariant\footref{supplemental}.
\begin{figure}[t]
  \centering
  \vskip -0.2in
  \includegraphics[width=0.245\textwidth]{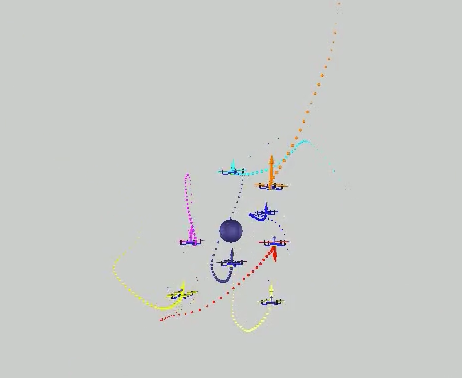}
  \includegraphics[width=0.245\textwidth]{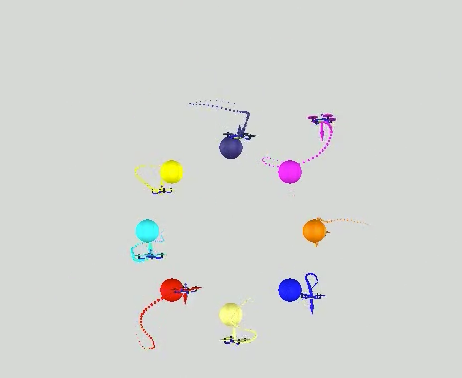}
  \includegraphics[width=0.245\textwidth]{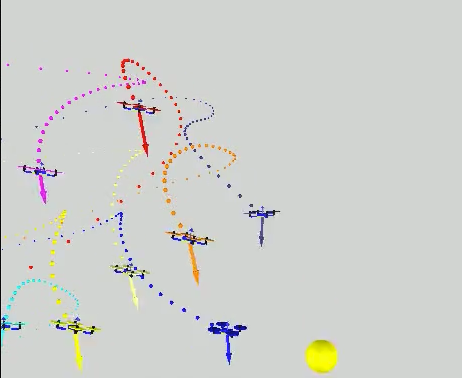}
  \includegraphics[width=0.245\textwidth]{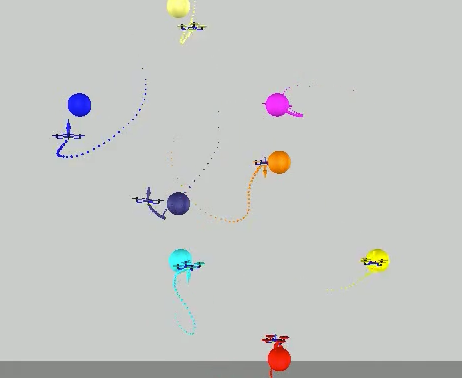}
  \vspace*{-5mm}
  \caption{Instances of the swarm in various scenarios.}\label{fig:scenarios}
  \vspace*{-4mm}
\end{figure}


\section{Experiments} \label{section:experiments}

To evaluate the effectiveness of incorporating artificial symmetries in MARL, we offer extensive experimentation on formation flight of swarms of $N$ Crazyflie quadrotors. Quadrotors have few inherent geometric symmetries to exploit (\cite{10156379,equivariant_downwash,10054413,HAMPSEY2023132}), so equivariant MARL approaches like \cite{chen2023rm,yu2023equivariant} are not applicable. Our scheme, presented in sections \ref{section:ArtificialEquivariance}-\ref{section:LieGraphormer}, allows for the embedding of artificial $SE(3)$ symmetry in the distributed controllers, even though the quadrotor model is not $SE(3)$-equivariant, as gravity is $O(3)$-invariant. The multi-agent environment implementation is based on \cite{huang2023quadswarm}, extended to include aerodynamic effects like drag and downwash, similarly to \cite{panerati2021learning}. For $x_i^d(t)\in \mathbb{R}^3$ desired target position, the state of each robot in the world-frame is $s_i(t) = (x_i(t),\dot{x}_i(t),R_i(t),\omega_i(t), x_i^d(t))$, where $R_i(t)\in SO(3)$ is the rotation matrix from the body frame to the world frame, $x_i(t)\in \mathbb{R}^3$ and $\omega_i(t)\in \mathbb{R}^3$ the position and angular acceleration in the world frame. Each robot receives neighborhood observations $o_i(t)=\cup_{j \in \mathcal{N}_i}s_j(t)$. The task is to learn a distributed, collision-free control policy that guides the swarm to a desired formation $x_i^d \in \mathbb{R}^{3 \times N},\forall i \in I_N$, i.e. learn  $\pi_\theta(u_i(t)|s_i(t),o_i(t))$ mapping state-observations to Gaussian distribution parametrized continuous actions $u_i(t)$. For enhanced generalization, a pool of geometric formations are used to construct diverse scenarios (static and dynamic formations, evasion pursuit etc.), e.g. Figure \ref{fig:scenarios}. The reward function for every quadrotor is $\Bar{R}(s_i(t),o_i(t),u_i(t)) = \Bar{R}^x +\Bar{R}^c + \Bar{R}^s$, where $\Bar{R}^x = -c_1 ||x_i(t)-x_i^d(t)||$ a penalty motivating the quadrotor to approach the target, $\Bar{R}^c=-c_2 \mathbf{1}_{||x_i(t)-x_j(t)||_2<d_{m}\,\forall j\in \mathcal{N}_i} -c_3 \sum_{j\in \mathcal{N}_i}[1-||x_i(t)-x_j(t)||_2/d_p]_+$ a collision penalty and $\Bar{R}^s = - c_4 ||\omega_i(t)||_2 - c_5 ||u_i(t)||_2$ an auxiliary reward facilitating learning relatively stable controllers at the early stages of training.

\begin{figure}[t]
    \centerfloat
    \vskip -0.2in
    \includegraphics[trim={0 25 0 0},clip,width=1.22\textwidth, angle=0]{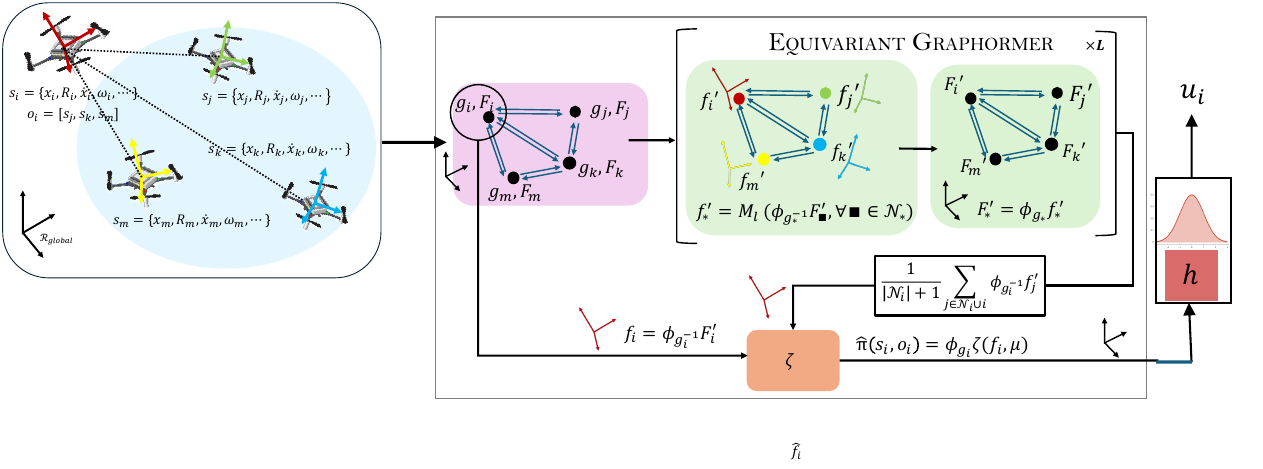}
    \vspace*{-3mm}
    \caption{Architecture schematic.} \label{figure:schematic}
    \vspace*{-3mm}
\end{figure}

\noindent\textbf{Architecture \& Training}:
If the neighborhood $\mathcal{N}_i$ is constructed via a Euclidean relative distance, for $SE(3)$ group actions Assumption \ref{assumption:G-invarinat neighborhood} stands. From $(s_i(t),o_i(t))$ every agent constructs a local approximation $\mathcal{G}_{i,t}$ of the graph representation $\mathcal{G}_{t}$ from Section \ref{intro:graph_representation}. As the rewards function is invariant to actions from $SE(3)$ and the $S_N$ permutation group, via Section \ref{section:ArtificialEquivariance} the policy becomes $\pi_\theta(u_i(t)|s_i(t),o_i(t)) = h_{\theta_p}(u_i(t)|\hat{\pi}_{\theta_e}(s_i(t),o_i(t)))$, where $h_{\theta_p}$ is parametrized by a simple MLP and $\hat{\pi}_{\theta_e}(s_i(t),o_i(t)) = \phi_{g_i} \zeta_{\theta_\zeta}(\phi_{g_i^{-1}} s_i(t) \oplus \frac{1}{|\mathcal{N}_i|+1}\sum_{j\in \mathcal{N}_i\cup \{i\}} \phi_{g_i^{-1}}f_{j}^L)$ the equivariant part of the policy with $\zeta_{\theta_\zeta}$ an MLP and $\{f_j^L,\forall j \in \mathcal{N}_i \cup \{i\}\} = M_{\theta_M}(\mathcal{G}_{i,t} \leftarrow s_i(t),o_i(t))$ the updated node features from the Group Equivariant Graphormer of Section \ref{section:LieGraphormer}. A schematic of our architecture is depicted in Figure \ref{figure:schematic}.
To train the distributed policy we use anasynchronous adaptation of the PPO algorithm from \cite{petrenko2020sf}. We use as baselines K-Attention and DeepSets \cite{batra21corl}, rMAPPO \cite{yu2022the} and InfoMARL \cite{10.5555/3618408.3619482}. Further details regarding architecture, hyperparameters, the reward function and training algorithm are offered in the supplemental material\footref{supplemental}. We, also, offer an adaptation of K-Attention with $SE(3)$ symmetry, following Lemma \ref{lem:canonic}. It should be noted that if the manifold or part of it forms a torsor shared by various groups and the reward function is invariant to actions induced by them, the problem admits multiple extrinsic symmetries. The selection process of the group if multiple choices are valid is outside the scope of this paper and is left as future work.
 
\noindent\textbf{Generalization}:
Table \ref{table:results_8} summarizes the evaluation of the policies trained in all scenarios with a swarm comprising of 8 quadrotors. The metrics were averaged over 50 episodes per task and include the average collected reward per agent, the average distance to the targets, the average number of collisions with environment or other agents, the success rate (drones remaining within a small distance of their assigned target while avoiding collisions) and the inter-agent collision rate. The policies with embedded $SE(3)$ symmetry, notably the $SE(3)$-K-Attention, outperform the baselines across all scenarios, attaining increased rewards and leading to fewer collisions and increased success probability. The average collision rate of the $SE(3)$-Graphormer across tasks is $2.5\%$ and $4.2\%$ for $SE(3)$-K-Attention compared to $7.4-16.1\%$ for the baselines. DeepSets and InfoMARL are more aggresive, occasionally achieving better positional rewards, as observed by the average distance to target, at the expense of collision related rewards. The reduced collisions of the symmetry-enhanced policies can be explained by the fact that during training the collision instances are significantly fewer than instances without them where policies mainly optimize the position reward, meaning that even thought there are enough samples for the policy to learn target tracking, the same is not true for collision avoidance. The symmetry-enhanced policies exploit the structure of the collision avoidance specification to shrink the hypothesis class and, thus, effectively restricting the models and leading to sampling efficient and accurate learning of collision avoidance without loss of expressivity of the final policy. Overall, the $SE(3)$-enhanced policies successfully manage to guide the quadrotors to the targets with significantly fewer collisions as indicated by the increased success rate.

\begin{table}[t!]
  \centerfloat
\begin{adjustbox}{width=1.25\textwidth}
\begin{tabular}{c}
     \textsc{Scenarios}\\

\begin{tabular}{c|ccccc|ccccc|ccccc|ccccc|} 
    \cline{2-21}
    & \multicolumn{5}{|c|}{\textbf{Static same goal}} & \multicolumn{5}{c|}{\textbf{Dynamic same goal}} & \multicolumn{5}{c|}{\textbf{Evasion Pursuit (Lissajous)}}& \multicolumn{5}{c|}{\textbf{Swarm-vs-Swarm}} \\
    \cline{2-21}
    & Rew & Dist & Col & Suc \% & IColR & Rew & Dist & Col & Suc \% & IColR & Rew & Dist & Col & Suc \% & IColR & Rew & Dist & Col & Suc \% & IColR \\
    \hline\hline
    K-Attention   & -6.49 & 0.56 & 1.25 & 71.9 & 0.14 
                    &  -7.28 & 0.77 & 1.5 & 65.6 & 0.17 
                    & -8.11 & 1.25 & 1.88 & 63.5 & 0.203
                    & -5.16 & 0.83 & 1 & 78.1 & 0.218\\
    rMAPPO        & -7.86 & 0.70 & 2.13 & 51.6 & 0.21 
                    & -8.54 & 1.49 & 1.75 & 53.1 & 0.205
                    & -8.65 & 1.59 & 1 & 71.9 & 0.125
                    & -8.13 & 0.71 & 2 & 46.8 & 0.235\\
    InfoMARL      & -2.73 & 0.42 & 0.05 & 98.8 & 0.013
                    &  -4.38 & 0.65 & 0.5 & 89.4 & 0.106 
                    & -5.68 & 0.83 & 0.68 & 85.1 & 0.148
                    & -4.15 & \textbf{0.50} & 0.71 & 83.3 & 0.167\\
    K-DeepSets      & -3.24 & 0.44 & 0.1 & 96.9 & 0.006  
                    &  -4.43 & 0.69 & 0.187 & 96.1 & 0.039  
                    & -5.05 & \textbf{0.69} & 0.51 & 88.1 & 0.117
                    &  -4.18 & 0.60 & 0.46 & 89.2 & 0.107\\
    SE(3)-K-Attention (Ours)  & -2.44 & 0.33 & \textbf{0} & \textbf{100} & \textbf{0} &
                        \textbf{-3.91 }& 0.87 & \textbf{0} & \textbf{100} & \textbf{0}
                        & -4.87 & 0.86 & \textbf{0.13} & 95.3 &\textbf{0.031}
                        &  -3.99 & 0.53 & 0.55 & 87.7 & 0.123\\
    SE(3)-Graphormer (Ours) &  \textbf{-2.41} & \textbf{0.32} & \textbf{0} & \textbf{100} &                 \textbf{0} 
                        &-4.21 & \textbf{0.64} & 0.125 & 96.8 & 0.031 
                        & \textbf{-4.53} & 0.74 & \textbf{0.13} & \textbf{96.8} & \textbf{0.031}
                        &  \textbf{-3.47} & 0.52 & \textbf{0.33} & \textbf{92.7} &  \textbf{0.073} \\
    \hline

    & \multicolumn{5}{|c|}{\textbf{Static different goals}} & \multicolumn{5}{c|}{\textbf{Dynamic different goals}} & \multicolumn{5}{c|}{\textbf{Evasion Pursuit (Bezier)}}& \multicolumn{5}{c|}{\textbf{Swap goals}} \\
    \cline{2-21}
    & Rew & Dist & Col & Suc \% & IColR & Rew & Dist & Col & Suc \% & IColR & Rew & Dist & Col & Suc \% & IColR & Rew & Dist & Col & Suc \% & IColR \\
    \hline
    K-Attention  & -3.88 & 0.41 & 0.5 & 87.5 & 0.063 &  
                    -5.37 & 0.74 & 0.92 & 80.2 & 0.099 
                    & -11.52 & 1.35 & 3 & 31.2 & 0.335
                    &  -3.91 & 0.57 & 0.50 & 88.6 & 0.113\\
    rMAPPO      & -3.92 & 0.55 & 0.33 & 80.2 & 0.04 &  
                    -6.33 & 1.01 & 1.17 & 74.4 & 0.119 
                    &-11.8 & 2.23 & 1.08 & 45.8 & 0.225 
                    & -4.01& 0.81 & 0.5 & 81.2 & 0.126\\
    InfoMARL    & -2.29 & \textbf{0.18} & 0.25 & 94.3 & 0.057 &
                    -3.97 & \textbf{0.60} & 0.42 & 90.5 & 0.104 
                    & -6.73 & 1.18 & 0.84 & 78.9 & 0.183
                    &-3.09 & 0.41 & 0.47 & 88.7 & 0.113\\
    K-DeepSets    & -2.73 & 0.32 & 0.07 & 98.1 & 0.017 & 
                    -4.43 & 0.72 & 0.38 & 90.9 & 0.091 
                    & -6.14 & 1.03 & 0.625 & 86.7 & 0.129
                    &-3.46 & 0.42 & 0.375 & 91.4 & 0.086\\
    SE(3)-K-Attention (Ours)   &  -1.91 & 0.24 & 0.03 & 99.2 & 0.007 
                                & -3.88 & 0.82 & 0.23 & 94.6 & 0.054 
                                & -6.71 & 1.46 & 0.37 & 80.2 & 0.082
                                &  \textbf{-2.71} & \textbf{0.40} & 0.15 & 96.2 & 0.038 \\
    SE(3)-Graphormer (Ours)   & \textbf{-1.87} & 0.21 & \textbf{0.02} & \textbf{99.3} &                                                 \textbf{0.003}&                                                                 \textbf{-3.18} & 0.81 & \textbf{0} & \textbf{100}                                              &\textbf{0}
                             & \textbf{-5.63} & \textbf{0.97} & \textbf{0.125} & \textbf{89.0} & \textbf{0.031}
                             & -2.82 & 0.55 & \textbf{0.125} & \textbf{96.9} & \textbf{0.031}\\

    \hline
\end{tabular}

\end{tabular}
  \end{adjustbox}
\caption{Evaluation for a swarm of 8 quadrotors in various scenarios.}\vspace*{-4mm}
\label{table:results_8}
\end{table}

\noindent\textbf{Scalability Ablation Studies}:
We examine the impact of extrinsic symmetries on scalability, i.e. using pretrained policy to control swarms of increasing size with zero-shot learning. The policies, trained for swarms of 8 quadrotors, are evaluated without further training in swarms of 16-128 quadrotors. The metrics are averaged over 50 episodes of static and dynamic scenarios (Table \ref{table:results_scalability}). As the room size remains unchanged, larger swarms lead to cluttered space, significantly increasing collisions affecting multiple drones in a cascade effect. However, the embedded symmetries leveraged locally are not affected by the size of the swarm. The symmetry-enhanced policies exhibit significantly fewer collisions than the baselines (for 128 drones the $SE(3)$-enhanced policies' collision rate rises to $\simeq16\%$ compared to $35-40\%$ for the baselines) as the swarm size progressively increases, while approaching reasonably the designated targets ($97\%$ and $84.64\%$ success rate in swarms of 64,128 compared to $<91\%$ and $<65\%$ for the baselines).

\begin{table}[h!]
  \centerfloat
\begin{adjustbox}{width=1.17\textwidth}
\begin{tabular}{c}
     \textsc{Swarm size}\\

\begin{tabular}{c|cccc|cccc|cccc|cccc|} 
    \cline{2-17}
    &  & \textbf{16} & & & & \textbf{32} & & &  & \textbf{64} & & & & \textbf{128} & & \\
    \cline{2-17}
    & Rew & Col & Suc \% & IColR & Rew & Col & Suc \% & IColR & Rew & Col & Suc \% & IColR & Rew & Col & Suc \% & IColR \\
    \hline\hline
    K-Attention   &  -5.38 & 1.6 & 84.6 & 0.154 &  
                        -5.40 &  2.8 & 86.6 & 0.134
                        & -5.58 &  7.6 & 82.8 & 0.172 
                        &  -7.70 &  69.2 & 58.9 & 0.406 \\
    rMAPPO &  -6.08 & 1.083 & 79.2 & 0.119 &
                -6.80 &  1.83 & 81.8 & 0.104 &  
                -7.79 & 16 & 57.2  & 0.352 &  
                -11.88 &  58.167 & 48.4 & 0.457 \\
    InfoMARL &  -4.07 &  0.343 & 96.7 & 0.033 & 
                    -4.72 & 2.333 & 88.02 & 0.119 & 
                    -5.19 & 11.25 & 77.3 & 0.227 &  
                    -7.03 &  40.29 & 64.6 & 0.351 \\
    K-DeepSets &  -\textbf{3.38}  & 0.2 & 97.5 & 0.025 & 
                    -3.92 & 1.875 & 90.2 & 0.098 & 
                    -4.56 & 3.8 & 90.2 & 0.121 & 
                    -10.51 &  163.17 & 64.9 & 0.348 \\
    SE(3)K-Attention (Ours)   &  -4.09 &  \textbf{0.067} & \textbf{99.2} & 0.008 & 
                                -4.41 & 0.583 & 96.6 & 0.034 & 
                                -4.96 & 2.43 & 93.7 & 0.063 &  
                                -5.66 & \textbf{12.75} &  83.77 & 0.16 \\
    SE(3)-Graphormer (Ours) &  -3.45 &  0.133 & 98.8 & \textbf{0.0125} & 
                                    \textbf{-3.69}  & \textbf{0.375} & \textbf{98.7} & \textbf{0.023}& \textbf{-4.29} & \textbf{0.82} & \textbf{97.5}  & \textbf{0.025} & 
                                    \textbf{-5.47} &  14.86 & \textbf{84.64} & \textbf{0.154}\\
    \hline
    
\end{tabular}

\end{tabular}
  \end{adjustbox}
\caption{Evaluation of zero-shot transferability to growing swarms. }
\label{table:results_scalability}\vspace*{-5mm}
\end{table}

\section{Conclusions \& Future Work}

This paper provides a novel methodology for leveraging extrinsic symmetries, indicated solely by the task, for systems without intrinsic ones and introduce the Group Equivariant Graphormer, an equivariant neural architecture adaptable to different symmetry groups. The experimental results of our work suggest that embedding extrinsic equivariance in MARL policies is beneficial for the generalization, scalability and sample efficiency. However, a task may be invariant to multiple symmetry groups that are compatible with the manifold but not exhibited by the dynamics. So how do we pick the symmetry group? We leave the symmetry selection strategy for future work.

\section{APPENDIX I - Proofs}\label{appendix:proofs}

\begin{proof}[Theorem \ref{theorem:equivariant_optimal_control}]
    From the invariance assumption of Definition \ref{eq:G-equivariant_problem1}, the value function of the centralized controller $u(t)=\pi(x(t))$ is equivariant under group actions induced by elements of $G$, i.e. $$V_{\psi_g(\pi)}(t,\phi_g(x(t)))=V_\pi (t,x(t)),\, \forall g \in G$$
Furthermore 
\begin{align*}
    V_\pi(t,x(t)) &=\int_{t}^{dt} \mathcal{L}(x(\tau),u(\tau)) d\tau +\int_{dt}^T \mathcal{L}(x(\tau),u(\tau)) d\tau\\
    &= \mathcal{L}(x(t),u(t))dt + V_\pi(t+dt,x(t+dt))+ O(dt)\\
    &= \mathcal{L}(x(t),u(t))dt + V_\pi(t,x(t))+ \nabla_t V_{\pi}(t,x(t)) dt +\\ &\qquad\qquad \nabla_{x(t)} V_{\pi}(t,x(t))^T \cdot \Tilde{f}(x(t),u(t)) dt + O(dt)\\
    \Rightarrow^{dt\rightarrow 0} -V_\pi(t,x(t)) &= \mathcal{L}(x(t),u(t))+ \nabla_{x(t)} V_{\pi}(t,x(t))^T \cdot \Tilde{f}(x(t),u(t))
\end{align*}
yields, for optimal value function $V^{*}$ acquired from the optimal control policy $u^{*}(t)= \pi^{*}(x(t))$, the Hamilton-Jacobi-Bellman equation
\begin{align}
\label{HJB}
    -V^{*}(t,x(t)) = \min_{u\in \Tilde{\mathcal{U}}} \big[\mathcal{L}(x(t),u(t))+ \nabla_{x(t)} V^{*}(t,x(t))^T \cdot \Tilde{f}(x(t),u(t)) \big]
\end{align}
for $\Tilde{\mathcal{U}} = \cup_{i\in I_N}\mathcal{U}$. The optimal control action
\begin{align}
\label{optimal action}
    \pi^{*}(x(t)) = \arg\min_{u\in \Tilde{\mathcal{U}}} \big[\mathcal{L}(x(t),u(t))+ \nabla_{x(t)} V^{*}(t,x(t))^T \cdot \Tilde{f}(x(t),u(t)) \big]
\end{align}
Under group actions induced by $g\in G$, the HJB optimal control (\ref{optimal action}) yields:
\begin{align}
\label{optimal_action_transformed}
    \pi^{*}(\phi_g(x(t))) &= \arg\min_{\Tilde{u}\in \Tilde{\mathcal{U}}} \big[\mathcal{L}(\phi_g(x(t)),\Tilde{u})+ \nabla_{\phi_g(x(t))} V^{*}(t,\phi_g(x(t)))^T \cdot \Tilde{f}(\phi_g(x(t)),\Tilde{u}) \big] \nonumber\\ \nonumber
    &= \arg\min_{\Tilde{u}\in \Tilde{\mathcal{U}}} \big[\mathcal{L}(x(t),\psi^{-1}_g(\Tilde{u}))+\nabla_{x(t)} V^{*}(t,x(t))^T \nabla_{\phi_g(x(t))}x(t)^T \cdot d\phi_g \Tilde{f}(x_i(t),\psi^{-1}_g(\Tilde{u}) \big]\\ \nonumber
    &= \arg\min_{\Tilde{u}\in \Tilde{\mathcal{U}}} \big[\mathcal{L}(x(t),\psi^{-1}_g(\Tilde{u}))+ \nabla_{x(t)} V^{*}(t,x(t))^T (d\phi_g^{-1} d\phi_g) \Tilde{f}(x_i(t),\psi^{-1}_g(\Tilde{u}) \big]\\ \nonumber
    &= \arg\min_{\psi_g(\Tilde{u})\in \Tilde{\mathcal{U}}} \big[\mathcal{L}(x(t),u)+ \nabla_{x(t)} V^{*}(t,x(t)) \Tilde{f}(x_i(t),u) \big]\\ 
    &= \psi_g(\pi^{*}(x(t)))
\end{align}
Since the system action is decomposed in individual agent actions on the graph via $\pi(x(t)) = \oplus_{i\in I_N} \pi_i(\cup_{j\in \mathcal{N}_i \cup \{i\}} x_j(t))$, then
\begin{align}\label{equation:control_decomposition}
    &\psi_g(\pi^{*}(x(t))) = \psi_g(\oplus_{i\in I_N} \pi_i^{*}(\cup_{j\in \mathcal{N}_i \cup \{i\}} x_j(t))) = \oplus_{i\in I_N} \psi_g(\pi_i^{*}(\cup_{j\in \mathcal{N}_i \cup \{i\}} x_j(t))) \nonumber\\
    &\pi^{*}(\phi_g(x(t))) = \oplus_{i\in I_N} \pi_i^{*}(\cup_{j\in \mathcal{N}_i \cup \{i\}} \phi_g(x_j(t))) \nonumber\\
    &\Rightarrow^{(\ref{optimal_action_transformed})} \pi_i^{*}(\cup_{j\in \mathcal{N}_i \cup \{i\}} \phi_g(x_j(t))) = \psi_g(\pi_i^{*}(\cup_{j\in \mathcal{N}_i \cup \{i\}} x_j(t))) 
\end{align}
i.e. the distributed controller must be equivariant w.r.t. actions induced by elements of the group $G$.

\end{proof}

\newlength{\arrow}
\settowidth{\arrow}{\scriptsize$10$}
\newcommand*{\mRightarrow}[1]{\xRightarrow{\mathmakebox[\arrow]{#1}}}
\begin{proof}[Theorem \ref{theorem:equivariant_input_extension}]
    By Lemma \ref{lemma:push_forward}, for the push-forward automorphism it stands that $d_{*}\phi_{g}\circ d_{*}\phi_{\Bar{g}}f(x,u) = d_{*}\phi_{g\Bar{g}}f(x,u)$. Thus:
    \begin{align*}
        d_{*}\phi_{\Bar{g}}F(x,\hat{u}) &= d_{*}\phi_{\Bar{g}} \circ \sum_k \alpha_k d_{*}\phi_{g_k}f(x,u_k)  = \sum_k \alpha_k d_{*}\phi_{\Bar{g} g_k}f(x,u_k) = F(x,w) \\
        \mathrel{\mathop{\Rightarrow}^{w=\psi_{\Bar{g}}(\hat{u})}_{x=\phi_{\Bar{g}}(\Bar{x})}} \quad& d\phi_{\Bar{g}}F(\phi_{\Bar{g}^{-1}}(\phi_{\Bar{g}}(\Bar{x})),\hat{u}) = F(\phi_{\Bar{g}}(\Bar{x}),\psi_{\Bar{g}}(\hat{u})) \\
        \Rightarrow & \quad d\phi_{\Bar{g}}F(\Bar{x},\hat{u}) = F(\phi_{\Bar{g}}(\Bar{x}),\psi_{\Bar{g}}(\hat{u}))
    \end{align*}
\end{proof}

\begin{proof}[Theorem \ref{theorem:equivariant_problem}]
    The optimal control solution of problem (\ref{eq:prob}) from the HJB equation is 
    \begin{align*}
        u^{*}_{I_N}(t)=&\argmin_{u_i \in \mathcal{U},i\in I_N} [\mathcal{L}(\oplus_{I_N}x_i(t) , \oplus_{I_N} h_\mathcal{U}(\hat{u}_i(t))) + <\nabla_{x(t)}V^{*}(t,\oplus_{I_N} x_i(t)),\oplus_{I_N}f(x_i(t),u_i(t)>] \\
    \stackrel{Remark (\ref{remark:2})}{=}&\argmin_{h_\mathcal{U}(\hat{u}_i) \in \mathcal{U},i\in I_N} [\mathcal{L}(\oplus_{I_N}x_i(t) , \oplus_{I_N} h_\mathcal{U}(\hat{u}_i(t))) + <\nabla_{x(t)}V^{*}(t,\oplus_{I_N} x_i(t)),\oplus_{I_N}F(x_i(t),h_\mathcal{U}(\hat{u}_i(t))>] \\
    =&h_\mathcal{U}(\argmin_{\hat{u}_i \in \hat{\mathcal{U}},i\in I_N} [\mathcal{L}(\oplus_{I_N}x_i(t) , \oplus_{I_N} h_\mathcal{U}(\hat{u}_i(t))) + <\nabla_{x(t)}V^{*}(t,\oplus_{I_N} x_i(t)),\oplus_{I_N}F(x_i(t),h_\mathcal{U}(\hat{u}_i(t))>] )\\
    =&h_\mathcal{U}( \hat{u}^{*}_{I_N}(t))
    \end{align*}
    Furthermore, as the associated dynamical system is G-equivariant by construction (see theorem \ref{theorem:equivariant_input_extension}), then, for $g\in G, \forall i\in I_N$, $h_\mathcal{U}(h_\mathcal{U}(\hat{u}))=h_\mathcal{U}(\hat{u})\,,\forall \hat{u}\in \hat{\mathcal{U}}$ and via the machinery of theorem \ref{theorem:equivariant_optimal_control} :
    \begin{align*}
        u_{I_N}^{*} &= h_\mathcal{U}(\hat{\pi}^{*}(\phi_g(\oplus x_i(t)))) = h_\mathcal{U}(\argmin_{\hat{u}_i \in \hat{\mathcal{U}},i\in I_N} [\mathcal{L}(\oplus_{I_N} \phi_g(x_i(t)) , \oplus_{I_N} h_\mathcal{U}(\hat{u}_i(t))) \\ &+ <\nabla_{\phi_g(x(t))}V^{*}(t,\oplus_{I_N} \phi_g(x_i(t))),\oplus_{I_N}F(\phi_g(x_i(t)),\psi_g(\psi_{g^{-1}}(h_\mathcal{U}(\hat{u}_i(t))))>] ) \\
        =&h_\mathcal{U}(\argmin_{\hat{u}_i \in \hat{\mathcal{U}},i\in I_N} [\mathcal{L}(\oplus_{I_N}x_i(t) , \oplus_{I_N} \psi_{g^{-1}} (h_\mathcal{U}(\hat{u}_i(t)))) \\
        &+ <d\phi_g^{-1}\nabla_{x(t)}V^{*}(t,\oplus_{I_N} x_i(t)),\oplus_{I_N}d\phi_g F(x_i(t),\psi_g^{-1}h_\mathcal{U}(\hat{u}_i(t))>] )\\
        =&h_\mathcal{U}(\argmin_{\hat{u}_i \in \hat{\mathcal{U}},i\in I_N} [\mathcal{L}(\oplus_{I_N}x_i(t) , \oplus_{I_N} h_\mathcal{U}(\psi_{g^{-1}}(h_\mathcal{U}(\hat{u}_i(t))))) \\
        &+ <\nabla_{x(t)}V^{*}(t,\oplus_{I_N} x_i(t)),\oplus_{I_N} F(x_i(t),h_\mathcal{U}(\psi_{g^{-1}}(h_\mathcal{U}(\hat{u}_i(t))))>] ) \\
        =&h_\mathcal{U}(\argmin_{h_\mathcal{U}(\hat{u}_i) \in \mathcal{U},i\in I_N} [\mathcal{L}(\oplus_{I_N}x_i(t) , \oplus_{I_N} h_\mathcal{U}(\psi_{g^{-1}}(h_\mathcal{U}(\hat{u}_i(t))))) \\
        &+ <\nabla_{x(t)}V^{*}(t,\oplus_{I_N} x_i(t)),\oplus_{I_N} F(x_i(t),h_\mathcal{U}(\psi_g^{-1}(h_\mathcal{U}(\hat{u}_i(t))))>])  \\
        =&h_\mathcal{U}(\psi_g((\argmin_{ \hat{u}_i \in \hat{\mathcal{U}},i\in I_N} [\mathcal{L}(\oplus_{I_N}x_i(t) , \oplus_{I_N} h_\mathcal{U}(\hat{u}_i(t))) \\
        &+ <\nabla_{x(t)}V^{*}(t,\oplus_{I_N} x_i(t)),\oplus_{I_N} F(x_i(t),h_\mathcal{U}(\hat{u}_i(t)))>])) =  h_\mathcal{U}(\psi_g( \hat{\pi}^{*}(\oplus x_i(t))))
    \end{align*}
    Therefore, there exists a G-equivariant lifted optimal control policy $\hat{\pi}^{*}: \mathcal{X}^N \rightarrow \hat{\mathcal{U}}$ such that $\pi^{*}(x)= h_\mathcal{U} \circ \hat{\pi}^{*} (x)$. Following the decomposition of the optimal control to that of indivisual agents, seen in equation \ref{equation:control_decomposition} from the proof of Theorem \ref{theorem:equivariant_optimal_control}, it yields $\pi^{*}(\cup_{j \in \mathcal{N}_i\cup\{i\}} \phi_g(x_j(t))) = h_\mathcal{U} \circ \hat{\pi}^{*}(\cup_{j \in \mathcal{N}_i\cup\{i\}} \phi_g(x_j(t)))= h_\mathcal{U} \circ \psi_g(\hat{\pi}^{*}(\cup_{j \in \mathcal{N}_i\cup\{i\}} x_j(t)))$.
\end{proof}
\begin{proof}[Lemma \ref{lem:canonic}]
    Assume that $f:G\times X\to Y$ is an equivariant function, then by definition we have that for all $q\in G$ and $p=(g,x)\in G\times X$:
    \begin{align*}
        f(\phi'_q[p])=\psi_q[f(p)]&\implies f(qg,\phi_q[x])=\psi_q[f(g,x)]
    \end{align*}
    then by setting $q=g^{-1}$ we get that:
    \begin{align*}
        f(g^{-1}g,\phi_{g^{-1}}[x])=\psi_{g^{-1}}[f(g,x)] &\implies f(e,\phi_{g^{-1}}[x])=\psi_{g^{-1}}[f(g,x)]\\
        & \implies \psi_g\left[f(e,\phi_{g^{-1}}[x])\right]=f(g,x)
    \end{align*}
so $f(g,x)=T_g\left[h(L_{g^{-1}}[x])\right]$ with $h(x)=f(e,x)$.
\par Additionally we can easily show that any function $f(g,x)=\psi_g[h(\phi_{g^{-1}}[x])]$ is equivariant since for any $p=(g,x)\in G\times X$, $q\in G$:
\begin{align*}
    f(\phi'_q[p])=f(qg,\phi_q[x])&=\psi_{qg}\left[h(\phi_{g^{-1}q^{-1}}\circ \phi_q[x])\right]=\psi_{qg}\left[h(\phi_{g^{-1}}[x])\right]\\
    &=\psi_{q}\left[\psi_g\left[h(\phi_{g^{-1}}[x])\right]\right]=\psi_{q}[f(g,x)]=\psi_q[f(p)]
\end{align*}
\end{proof}

\section{APPENDIX II - Literature Review}\label{appendix:Litterature_review}

\noindent\textbf{Symmetries in RL \& Control}: Exploiting symmetries in MDPs has been introduced in \cite{Ravindran2001SymmetriesAM,Zinkevich2001SymmetryIM}. Recent work focus on symmetries for deep RL as a data-efficient approach via embedding symmetries in policy networks (\cite{vanderpol2020plannableapproximationsmdphomomorphisms,NEURIPS2020_2be5f9c2,zhu2022sampleefficientgrasplearning,simm2020symmetryawareactorcritic3dmolecular}), invariant data augmentation (\cite{NEURIPS2020_e615c82a,kostrikov2021imageaugmentationneedregularizing}), symmetric filters (\cite{pmlr-v37-clark15}), equivariant trajectory augmentation (\cite{pmlr-v162-weissenbacher22a,9158366,mavalankar2020goal}). \cite{NEURIPS2020_2be5f9c2} proposed an equivariant architecture on $C_n$ symmetries of discrete rotations and \cite{pmlr-v164-wang22j,wang2020policylearningse3action} extended it to continuous $SO(2)$ and $SE(2)$ symmetries. \cite{mondal2020groupequivariantdeepreinforcement} applied an $E(2)$-equivariant network to Q-learning in an Atari game domain. \cite{Yu2022EquivariantRL}  constructed an $S^1$ symmetry for a quadrotor policy. \cite{sonmez2024exploitingsymmetrydynamicsmodelbased} used Cartan’s moving frame method to construct equivalent states and invariant input features for the policy, independent of the reward function. The effectiveness of learning equivariant policies, with continuous or discrete symmetries, for a single agent has been demonstrated in symmetric locomotion (\cite{su2024leveraging,abdolhosseini2019learning,ordonez2023discrete}), planning \cite{zhao2023integrating}, grasping and manipulation (\cite{zhu2022sample,10160728,pmlr-v164-wang22j,nguyen2023equivariant}). Unlike the aforementioned that assume apriori knowledge of symmetries, \cite{pmlr-v162-weissenbacher22a} infers symmetries from data and augments the training data using these symmetries.
Symmetries have been, also, utilized in control and state estimation (\cite{1103421,MAHONY2013617,mahony2020equivariantfilterdesignkinematic,mahony2020equivariantsystemstheoryobserver}), dynamic programming \cite{maidens2018symmetryreductiondynamicprogramming} and reachability analysis \cite{8277172} to reduce computational complexity. 
However, these approaches are only suitable for single agent or centralized
multi-agent controllers for system dynamics and tasks with inherent symmetries, thus severely limiting their applicability. We solve the problem of leveraging symmetries, even in systems that do not exhibit them, while still allowing distributed execution.

\noindent\textbf{Permutation symmetries in multi-agent systems}:
Permutation invariance is another symmetry often appearing in homogeneous MARL where the dynamics and reward functions of all agents are the same, so identities are disregarded and any indexing permutation of them would lead to the same indexing permutation of the optimal actions as only the observations are used. This property directly limits the required sample complexity for training the distributed policy, and can be usually exploited via policy sharing (\cite{chen2022communicationefficient,NIPS2017_c2ba1bc5}), permutation equivariant critics (\cite{Almasan_2022,liu2019picpermutationinvariantcritic,rashid2020monotonicvaluefunctionfactorisation,liu2019picpermutationinvariantcritic}) or mean field approximation (\cite{pmlr-v80-yang18d}). For policy sharing, a common approach is to use coordination graphs to explicitly formulate symmetries and learn local Q-functions. \cite{pmlr-v119-boehmer20a} factored the joint value function according to a coordination graph into payoffs between pairs of agents while \cite{Pol2016Coordinated} considered the deconstruction of the problem as a linear combination of local subproblems. Another approach is to use graph-based networks that are inherently permutation equivariant, like \cite{jiang2020graphconvolutionalreinforcementlearning}. \cite{pmlr-v119-hu20a} used known permutation symmetries to improve zero-shot coordination in game that require coordinated symmetry breaking. Though, permutation equivariant methods exploit the homogeneity structure, they ignore the geometry of the problem. We leverage both by introducing a group equivariant graph neural network.

\noindent\textbf{Geometric Symmetries in MARL}: From a geometric standpoint, few symmetries in MARL have been explored. \cite{vanderpol2022multiagentmdphomomorphicnetworks} introduced Multi-Agent MDP Homomorphic Networks to utilize $C_n$ symmetries arising between different configurations of the agents and their local observations, with, however, the input size of the proposed network that scaling with $|C_n|$ and Euclidean equivariance only preserved for discrete rotations. \cite{pmlr-v202-chen23i} inject an $E(3)$ symmetry in subequivariant morphology-agnostic RL policies and Q-function such that the policy can generalize to all directions, improving exploration efficiency.
\cite{chen2023rm} characterize a subclass of Markov games with a general notion of symmetries that admits the existence of symmetric optimal values-policies and embed $E(3)$ constraints in graph-based multi-agent actor-critic methods. \cite{mcclellan2024boostingsampleefficiencygeneralization} used an adaptation of EGNNs to demonstrated how $O(n)$ equivariance boosts sample efficiency and generalization in MARL. All of these works use equivarinace-by-construction policies, usually based on equivariant operations on message-passing networks (\cite{pmlr-v139-satorras21a,pmlr-v139-schutt21a,bekkers2024fast}), lifting to the lie algebra (\cite{finzi2020generalizing,mironenco2024lie,macdonald2022enablingequivariancearbitrarylie}), group infinitesimal generators (\cite{NEURIPS2021_148148d6}), gauge theory (\cite{NEURIPS2021_e57c6b95,dehaan2021gaugeequivariantmeshcnns}), vectorized features (\cite{Deng2021VectorNA}) or steerability via harmonic analysis (\cite{thomas2018tensor,fuchs2020se3transformers}) This assumes knowledge of a predefined geometric symmetry of the problem and, thus, are not transferable to MARL problems with different symmetries. We propose a simple but modular Group Equivariant Graphormer that can be adapted for a variety of symmetries in differently structured MARL problems via canonicalizing group actions on tensorial graph features. Furthermore, all of these works attempt to exploit only existing symmetries in the systems, whereas the current paper extends to systems with partial or broken symmetries via embedding them in associated artificial symmetries-enhanced systems to include desirable equivariant properties.

\section{APPENDIX III - Experiment Details}\label{appendix:hyperparams}
\begin{figure}[t!]
  \centering
  \includegraphics[width=0.245\textwidth]{files/figures/Screenshot1.png}
  \includegraphics[width=0.245\textwidth]{files/figures/Screenshot2.png}
  \includegraphics[width=0.245\textwidth]{files/figures/Screenshot3.png}
  \includegraphics[width=0.245\textwidth]{files/figures/Screenshot4.png}
  \includegraphics[width=0.245\textwidth]{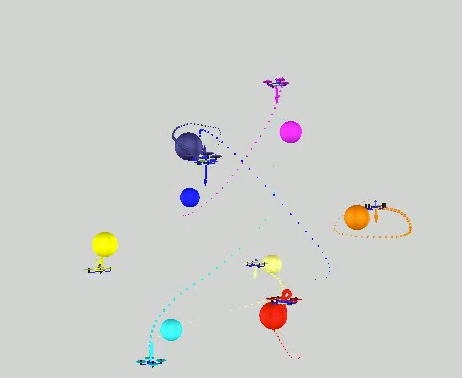}
  \includegraphics[width=0.245\textwidth]{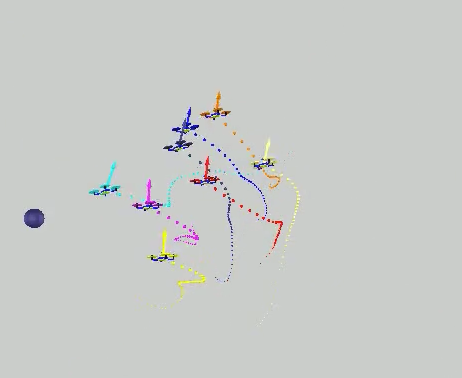}
  \includegraphics[width=0.245\textwidth]{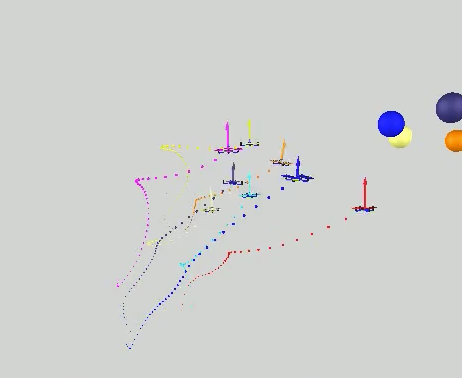}
  \includegraphics[width=0.245\textwidth]{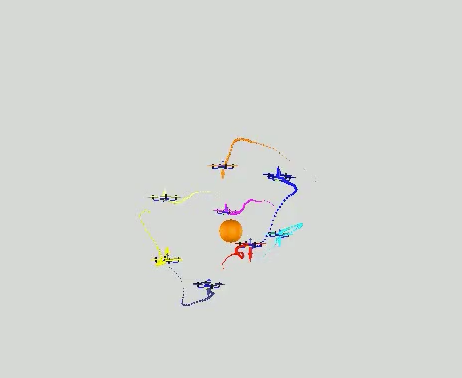}
  \caption{Instances of the swarm in various scenarios}
\end{figure}
\subsection{Architecture Details}
The $SE(3)$-Equivariant Graphormer consists of 3 layers with layer normalization. For the hidden and output layers we use fibers of types 0 and 1 with multiplicity 60 for a total hidden feature size of 240. The MLP $\zeta_{\theta_m}$ has size [(state size,256)(256,128),(128+mean embedding size,256),(256,256)] with tanh activations and the projection MLP $h_{\theta_p}$ has size [(256,256),(256,4)].

\subsection{Training Hyperparameters}
The physics simulation runs integrations at a 250Hz frequency and control at 100Hz. The parameters of the Craziflie 2.0 can be found in \cite{Frster2015SystemIO}.

\noindent\textbf{Reward function}:To construct a reward scheme that is independent of the control frequency of the system, reward parameters are multiplied by the control timestep $dt$. For each collision the penalty is only applied in the first instance that the collision was detected, meaning that the collision coefficient is independent of $dt$. That way actions taken under collision status are not penalized to learn recovering behaviors.

\begin{center}
\begin{tabular}{c|c|c}
    \hline
     \textbf{Coefficient name}& \textbf{Symbol}  & \textbf{Value} \\
     \hline
     Position penalty & $c_1$ & $0.5\cdot dt$\\
     Collision penalty & $c_2$ & $5$\\
     Smooth proximity & $c_3$ & $5\cdot dt$\\
     Angular velocity & $c_4$ & $0.01\cdot dt$\\
     Thrust penalty &  $c_5$ &  $0.05\cdot dt$\\
     \hline
\end{tabular}
\end{center}
\noindent\textbf{Training algorithm}:We use APPO implented in Sample Factory \cite{petrenko2020sf} with learning rate $10^{-4}$, discount coefficient $0.99$, Xavier uniform policy initialization, Adam optimizer with $\beta_1=0.9,\beta_2=0.995,\epsilon=2\cdot10^-6$, gradient norm clipping $5.0$, rollout length $128$, batch size $4096$ and $5$ training epochs for the PPO.


\section{APPENDIX IV - Group Theory}\label{appendix:group_theory}
For an extensive introduction on Lie Groups and manifolds the authors recommend \cite{gallier}.
\subsection{Group Representation Theory}
A group $(G,\cdot)$ is a set $G$ equipped with an operator $\cdot:G\times G \rightarrow G$ that satisfies:
\begin{enumerate}
    \item Closure: $\forall g,h \in G,\, g\cdot h \in G$
    \item Identity: $\exists e \in G$ such that $e \cdot g = g \cdot e = e$
    \item Associativity: $\forall g,h,f \in G,\, g\cdot (h \cdot f) = (g \cdot h) \cdot f$
    \item Inverse: $\forall g \in G,\, \exists g^{-1}$ such that $g^{-1} \cdot g = g \cdot g^{-1} = e$
\end{enumerate}
Note that groups are not necessarily commutative, i.e. $\exists (G,\cdot)$ such that $g \cdot h \neq h \cdot g, \forall g,h \in g$.  If the binary operation is commutative, the group is abelian. If the set $G$ is equipped with a topology with the binary operator and the inverse map continuous, $G$ is a topological group. If that topology is compact, the group $G$ is a compact group. If G is a smooth manifold and the binary operator and the inverse map are smooth, it is a Lie group. A subgroup $(H, ·)$ of a group $(G, ·)$ is a group such that $H \subseteq G$. Given a group $(G,\cdot)$, we denote the action on the space $X$ with $L_g:X \rightarrow X,\, \forall g \in G$. The transformation/operator satisfies the following properties:
\begin{enumerate}
    \item Trivial action: for $e\in G$ with $e \cdot g = g \cdot e = e$,   $L_e[g]=g,\, \forall g \in G$
    \item Closure: $\forall g,h,q \in G$, $L_g \circ L_h [q] = L_{g\cdot h}[q]$
\end{enumerate}
where $\circ$ is the composition of transforming actions. If $\forall x,y \in X,\, \exists g\in G$ such that $L_g[x]=y$, then $X$ is a homogeneous space for group $G$.
\begin{definition}
    A \textbf{\textit{representation}} $(G,\rho)$ of an algebraic group is a morphism $\rho:G \rightarrow GL(V)$ mapping linearly from the group $G$ to the general linear group $GL(V)$, for $V$ some vector space over some field, with the group action defined by the linear operator, i.e. $L_g[x]=\rho(g)x\,,\, \forall x \in V, g\in G$. 
\end{definition}
If the morphism $\rho(\cdot)$ is a group homeomorphism, i.e. $\rho(g\cdot h)= \rho(g) \cdot \rho(h)\,,\, \forall g,h \in G$, then the representation $(V,\rho)$ is a linear group representation.  The \textbf{\textit{degree}} of a linear representation is the dimension of the corresponding vector space $V$ .

\begin{definition}[\textbf{Equivariance}] Given group $G$ and the set of actions $L_g:X \rightarrow X\,,\, \forall g\in G$ and the the set of actions $T_g:Y \rightarrow Y\,,\, \forall g\in G$, the map $f:X \rightarrow Y$ is \textbf{$(G,L_*,T_*)$-equivariant} if $T_g[f(x)]=f(L_g[x])\quad \forall x\in X, g\in G$
\end{definition} 
\begin{definition}[\textbf{Invariance}]
Given actions $L_g:X \rightarrow X\,,\, \forall g\in G$, the map $f:X \rightarrow Y$ is \textbf{$(G,L_*,T_*)$-invariant} if $f(x)=f(L_g[x])\quad \forall x\in X, g\in G$
\end{definition} 
A \textbf{unitary} finite dimensional representation of a group $G$ is a representation of $G$ on a complex finite dimensional vector space $V$ over $\mathbb{C}$ equipped with a ${G}$-invariant positive definite Hermitian form (linear in the first argument, anti-linear in the second), i.e. $<\rho(g)v, \rho(g)w>=<v,w>\,,\,\forall g \in G,v,w \in V$.
If $V$ is an inner product space and the morphism $\rho$ is continuous and preserves the inner product, it is a unitary representation. 

If $f:X \rightarrow Y$ is linear and equivariant then it it an \textit{\textbf{intertwiner}} with respect to $(G,L_*,T_*)$.  If $\exists W \subset V$ such that $L_g[w]=(\rho(g)\cdot w) \in W\,,\, \forall w \in W, g\in G$, then $W$ is a $G$-invariant subspace of $V$ and the representation $(W,\rho)$ is a \textbf{\textit{subrepresentation}} of $(V,\rho)$.  

\begin{definition}
A representation $W$ is \textbf{\textit{irreducible}} if and only if ${0}$ and $W$ are the only subrepresentations. 
\end{definition}

If $\exists \tilde{W} \subset W, \tilde{W} \neq {0}$ subrepresentation of $W$, i.e. $W$ is reducible, then $\exists Q$ invertible matrix such that $\rho(g)=Q^{-1} [\oplus_{i} \rho_i(g) ] Q\,,\, \forall g\in G$. A linear representation $V$ is \textbf{completely reducible} if $V$ is a direct sum of irreducible representations.

Given representations $(G,\rho_V)\,,\, \rho_V:G \rightarrow GL(V)$ and  $(G,\rho_W)\,,\, \rho_W:G \rightarrow GL(W)$ with respective linear operator actions $L_g:V \rightarrow V, T_g: W \rightarrow W$, a linear morphism of representations $f:V \rightarrow W$ is a \textbf{$G$-map} if it is $(G,L,T)-$invariant, i.e. $f(L_g[v])=f(\rho_V(g)v)=\rho_W(g) f(v) = T_g[(f(v))]\,,\, \forall g\in G,v \in V$. If that linear morphism is invertible, the two representations of the group $G$, $(G,\rho_1),(G,\rho_2)$ are \textbf{\textit{equivalent}}, i.e. $\exists Q$ similarity matrix such that $\rho_1(g)= Q^{-1} \rho_2(g) Q\,,\, \forall g \in G$.  

\begin{lemma}[\textbf{Shur}]
    Let $(V,\rho_V)\,,\, \rho_V:G \rightarrow GL(V)$ and $(W,\rho_W)\,,\, \rho_W:G \rightarrow GL(W)$ be two irreducible representations of $G$  with actions $L_g:V \rightarrow V, T_g: W \rightarrow W$ and $f:V \rightarrow W$ a $G$-map over the respective vector spaces. Then, the morphism $f(\cdot)$ is either an isomorphism or $f(\cdot)=0$. For any two such isomorphisms $f_1:V \rightarrow W$, $f_2:V \rightarrow W$ it holds that $f_1=k f_2$.
\end{lemma}


For $(V,\rho_V)$ and $(W,\rho_W)$ irreducible representations of $G$, Shur's lemma means that: 
\begin{enumerate}
    \item If $V.W$ are not isomorphic spaces, then there exist no trivial intertwiners with respect to $(G,L_*,T_*)$ between them.
    \item If finite-dimensional spaces  $V\equiv W \subseteq \mathbb{C}$ and morphisms  $\rho_V=\rho_W$, then all intertwiners are scalar multiplicities of the identity.

\end{enumerate}

\begin{theorem}[\textbf{Krull-Schmidt}]
    Suppose V is a finite-dimensional representation of a finite group $G$. All decompositions $V = [\oplus_i V_i] $ into irreducible representations are unique up to isomorphism and reordering.
\end{theorem}

\begin{lemma}
    Let $G$ be a finite group of order coprime to the characteristic of a field $k$. If $V$ is a representation of $G$ and $W$ is a subrepresentation of $V$ , then there exists a subrepresentation $U$ of $V$ such that $V = W \oplus U$.
\end{lemma} 

\begin{theorem}[\textbf{Maschke}]
    If $G$ is a finite group with $(V,\rho)$ finite dimensional representation over $\mathbb{C}$, then every finite-dimensional regular representation of G is completely reducible  into as a direct sum of irreducible representations $V_i$, i.e. $V = [\oplus_i V_i] $ and $|G| = \sum_{i} dim(V_i)^2 $. Every finite-dimensional unitary representation of a compact group is a direct sum of unitary irreducible representations
\end{theorem}

\begin{lemma}
    Assuming representations $(G,\rho_V)\,,\, \rho_V:G \rightarrow GL(V)$ and  $(G,\rho_W)\,,\, \rho_W:G \rightarrow GL(W)$ with respective linear operator actions $L_g:V \rightarrow V, T_g: W \rightarrow W$, the map $\rho_{VW}:G \rightarrow V \otimes W$, obtained via the tensor product $\rho_{VW}(g)=\rho_{V}(g) \otimes \rho_W(g)\,,\, \forall g \in G$, is a valid representation of $V \otimes W$. 
\end{lemma} 

\begin{remark}
    Note that the above lemma agrees with the G-action one obtains from the canonical isomorphism $Hom_k(W,V) \cong W^{\vee}\otimes V $.
\end{remark}
\begin{definition}
    A representation  $(G,\rho_V)\,,\, \rho_V:G \rightarrow GL(V)$ is \textbf{faithful} if  $\rho_V$ is injective as a map of sets.
\end{definition}

\begin{lemma}
    Let $G$ be a finite group and $\rho:G \rightarrow GL(\mathbb{C},n)$ a faithful representation of $G$ Then every irreducible representation of $G$ is contained in some tensor power of $\rho$.
\end{lemma}

\noindent The \textbf{character} of the representation $(G,\rho)$ is a map ${\chi}_{\rho}:G \rightarrow \mathbb{C}$ such that ${\chi}_{\rho} (g)= tr(\rho(g))\,,\, \forall g\in G$. We define the inner product of characters $<\chi_i, \chi_j> = \int_{G} \chi_i(g) \overline{\chi_{j}(g)}dg$. Then, the following properties hold:
\begin{enumerate}
    \item If $(V,\rho_V),(W,\rho_W)$ isomorphic representations of group $G$, then $\chi_V=\chi_W$. The converse is only true for semisimple representations, which include unitary representations and all representations of finite or compact groups.
    \item The distinct characters of irreducible representations of compact groups are orthogonal, i.e. $<\chi_i, \chi_j>=0\,,\, \forall i \neq j$.
    \item A representation of a compact group is irreducible if and only if  $<\chi,\chi>=1$.
    \item The character of a representation that is a direct sum of representations, is the sum of the characters, i.e. for $V=[\oplus_{i\in K} V_i]$, $\chi_V = \sum_{i\in K} \chi_{V_i}$.
\end{enumerate}

Given a finite dimentional representation $(V,\rho)$ of $G$, the \textbf{\textit{restriction}}  $Res_{H}^{G}(V)$ to $H$ is the representation $(V,\rho|_H)$ where the underlying vector space is the same, but the group homomorphism $\rho|_H:H \rightarrow GL(V)$ has domain $H$. Now, suppose $H$ is a subgroup of a finite locally compact group $G$ and $W$ is a finite dimensional representation of $H$. The \textbf{\textit{induced representation}} is a representation $Ind_{H}^{G}(W)$ of $G$ together with an $H-$invariant map $p:W \rightarrow Ind_{H}^{G}(W)$ such that for any other representation $V$ of $G$ and $H-$invariant map $\phi:W \rightarrow V$, there exists unique $G-$equivariant map $\psi: Ind_{H}^{G}(W) \rightarrow V$ such that $\psi \circ p = \phi$. Induced representations exist and are unique up to unique isomorphism.

Let $\rho$ be a unitary representation of a compact group $G$. We denote $\Phi_{x,y}=<\rho(g)x,y>\,,\, g\in G$ the matrix coefficients of the representation $\rho$, where $\rho_{ij}(g) = \Phi_{e_i,e_j}(g)\,,\, g\in G$ for basis vectors $e_i,e_j$. 

\begin{theorem}[\textbf{Peter-Weyl}]
    Let $G$ be a compact group. 
\begin{enumerate}
    \item The linear span of the set of matrix coefficients of unitary irreducible representations of the group $G$ is dense in the space of continuous complex values functions on $G$, under uniform norm. 
    \item Let $\hat{G}$ be the set of equivalence classes of unitary irreducible representations  of $G$. For a unitary irreducible representation $\rho$ of $G$ on a complex Hilbert space, we denote its representation space by $H_\rho$, $dim H_\rho = d_\rho$, and its equivalence class by $[\rho]\in \hat{G}$. If $\pi$ is a reducible unitary representation of $G$, it splits in the orthogonal direct sum $H_\pi = \oplus_{[\rho] \in \hat{G}}M_\rho$, where $M_\rho$ is the largest subspace where $\pi$ is equivalent to $\rho$. Each $M_\rho$ splits in equivalent irreducible subspaces $M_\rho = \oplus_{i=1}^{n}H_\rho$, with $n$ the multiplicity of $[\rho]$ in $\pi$.
    \item Let $\epsilon_\rho$ be the linear span of the matrix coefficients of $\rho$ for $[\rho]\in \hat{G}$. Then, $L^2(G) = \oplus_{[\rho]\in \hat{G}}\epsilon_\rho$. If $\pi$ is a regular representation on $L^2(G)$, the multiplicity of $[\rho]\in \hat{G}$ in $\pi$ is $d_\rho$. An orthonormal basis of $L^2(G)$ is $\{ \sqrt{d_\rho}\rho_{ij}|i \geq 1, j \leq d_\rho, [\rho]\in \hat{G}\}$, $\rho_{ij}(g)= \Phi_{e_i,e_j}$. Constracting the basis requires choosing a representative per equivalence class.
\end{enumerate}
\end{theorem} 

The Peter-Weyl theorem gives an explicit orthonormal basis for $L^2(G)$, constructed from irreducible representations of a group $G$. The basis is formed by matrix coefficients



\acks{This work was supported by the ARL grant DCIST CRA W911NF-17-
2-0181. The authors thank Evangelos Chatzipantazis for the helpful discussions.}

\bibliography{bibliography}

\end{document}